\documentclass[10pt,twocolumn,letterpaper]{article}

 \usepackage[pagenumbers]{cvpr} % To force page numbers, e.g. for an arXiv version

\usepackage{blindtext}
\usepackage{graphics}
\usepackage{caption}
\usepackage{colortbl}
\definecolor{mygray}{RGB}{234,234,234}
\usepackage{tabularx}

\definecolor{cvprblue}{rgb}{0.21,0.49,0.74}
\usepackage[pagebackref,breaklinks,colorlinks,allcolors=cvprblue]{hyperref}

\def\paperID{0000} % *** Enter the Paper ID here
\def\confName{CVPR}
\def\confYear{2026}

\title{EmoStyle: Emotion-Driven Image Stylization}

\begin{document}
	
	\author{Jingyuan~Yang,
		Zihuan~Bai,
		Hui~Huang\footnotemark[1]\\
		CSSE, Shenzhen University \\	
		{\tt\small \{jingyuanyang.jyy, baizihuan23, hhzhiyan\}@gmail.com}
		\vspace{-25pt}
	}

\twocolumn[{
	\renewcommand\twocolumn[1][]{#1}
	\maketitle
	\begin{center}
		\centering
		\includegraphics[width=\linewidth]{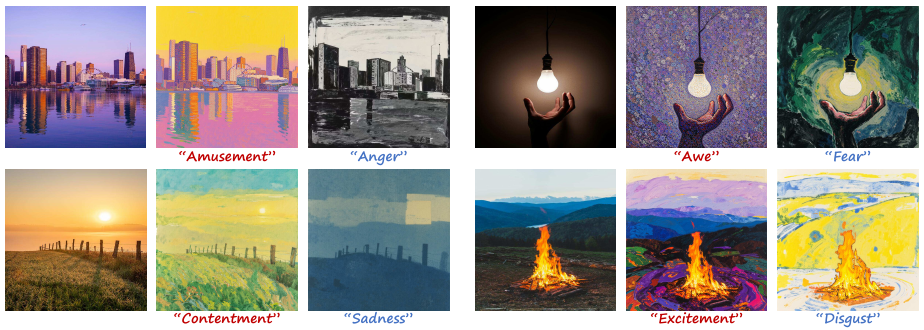}
		%		(a)\hspace{32mm}(b)\hspace{32mm}(c)\hspace{32mm}(d)\hspace{32mm}(e)
		%		\vspace{-10pt}
		\captionof{figure}{Affective Image Stylization with EmoStyle, aiming to transform user-provided images through artistic stylization to evoke specific emotional responses. It requires only emotion words as prompts, eliminating the need for reference images or detailed text descriptions.}
		%	\vspace{-10pt}
		\label{fig:teaser}
	\end{center}
}]

\begin{abstract}
	
Art has long been a profound medium for expressing emotions.
While existing image stylization methods effectively transform visual appearance, they often overlook the emotional impact carried by styles.
To bridge this gap, we introduce Affective Image Stylization (AIS), a task that applies artistic styles to evoke specific emotions while preserving content.
We present EmoStyle, a framework designed to address key challenges in AIS, including the lack of training data and the emotion–style mapping.
First, we construct EmoStyleSet, a content-emotion-stylized image triplet dataset derived from ArtEmis to support AIS.
We then propose an Emotion–Content Reasoner that adaptively integrates emotional cues with content to learn coherent style queries.
Given the discrete nature of artistic styles, we further develop a Style Quantizer that converts continuous style features into emotion-related codebook entries.
Extensive qualitative and quantitative evaluations, including user studies, demonstrate that EmoStyle enhances emotional expressiveness while maintaining content consistency.
Moreover, the learned emotion-aware style dictionary is adaptable to other generative tasks, highlighting its potential for broader applications.
Our work establishes a foundation for emotion-driven image stylization, expanding the creative potential of AI-generated art.
	
\end{abstract}    
\section{Introduction}
\label{sec:intro}

\begin{flushleft}
	\textit{``Art is not a handicraft, it is the transmission of feelings.''}
\end{flushleft}
\vspace{-15pt}
\begin{flushright}
	\textit{--Leo Tolstoy}
\end{flushright}
\vspace{-10pt}

Beyond aesthetic appreciation, art has long served as a powerful medium for conveying emotions, allowing artists to elicit deep emotional responses from viewers~\cite{joshi2011aesthetics, barwell1986does}.
For instance, while \textit{Van Gogh}'s \textit{Starry Night} evokes awe and wonder, \textit{Edvard Munch}'s \textit{The Scream} conveys fear and despair.
These expressive contrasts raise a fundamental question: \textit{Can we automatically create images that are not only visually compelling but also emotionally resonant?}

Style transfer (ST) applies artistic techniques to modify an image’s appearance while preserving its structure, which has advanced significantly in recent years~\cite{hertz2024style, chung2024style, wang2025omnistyle, xing2024csgo}.
However, most methods rely on reference images, limiting their applicability when no style image is available.
Text-guided style transfer~\cite{kwon2022clipstyler, liu2023name, huang2024diffstyler} offers greater flexibility by allowing users to specify styles through prompts.
Yet, current methods usually describe style via medium (\textit{oil painting}), artist (\textit{Claude Monet}), art movement (\textit{Cubism}), or iconic artworks (\textit{Mona Lisa}), necessitating a certain degree of artistic knowledge.
While not everyone is an artist, emotions are universally understood.

In Affective Computing~\cite{picard2000affective}, researchers have extensively studied how images convey emotions~\cite{zhao2024err, zhao2023toward, achlioptas2021artemis, bose2021understanding, chen2024exploring}. 
With the rise of generative AI, the focus has shifted from understanding emotions to Affective Image Manipulation (AIM), which modifies images to evoke target emotional responses~\cite{yang2025emoedit}.
Some methods alter emotional tones through color adjustments~\cite{yang2008automatic, wang2013affective, liu2018emotional}, while others modify visual content~\cite{yang2024emogen, yang2025emoedit, lin2025make}.
However, these techniques primarily focus on generating realistic images, overlooking artistic styles as expressive tools for emotional communication.
Recent studies have explored emotion-aware style transfer~\cite{sun2023msnet, weng2023affective}, yet they often rely on reference images~\cite{sun2023msnet} or require carefully crafted textual descriptions~\cite{weng2023affective}, limiting their practicality and accessibility.

\begin{figure}
	\centering
	\includegraphics[width=\linewidth]{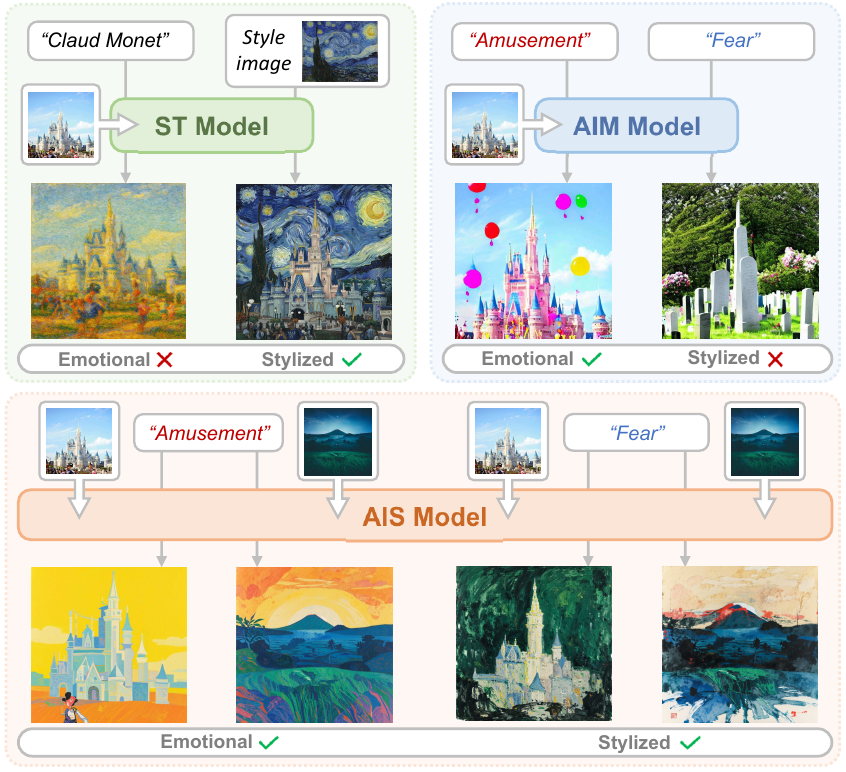}
	\vspace{-5pt}
	\caption{Comparison among Style Transfer (ST), Affective Image Manipulation (AIM), and our proposed Affective Image Stylization (AIS). While ST focuses on style and AIM emphasizes emotion, AIS generates emotionally expressive and stylized results.
	}
	%	\Description{}
	\label{fig:intro}
	\vspace{-5pt}
\end{figure}

Therefore, we introduce Affective Image Stylization (AIS), a new task that integrates artistic style into images to evoke specific emotional responses while preserving semantic content.
As shown in Fig.~\ref{fig:intro}, AIS bridges the gap between ST and AIM by combining the strengths of both: it evokes emotions like AIM while producing stylized results like ST, prompted only with emotion words.
AIS introduces two key challenges: (1) \textit{Lack of training data}: no dataset provides content-emotion-stylized image triplets for learning emotion-aware stylization; (2) \textit{Emotion-style mapping}: generating stylistic variations that remain content-consistent yet emotionally expressive is highly non-trivial.
We therefore introduce both a tailored dataset and a framework to address these challenges.

AIS requires the model to understand \textit{what} the image depicts (content) and \textit{which} emotion it should express, in order to determine \textit{how} it should appear (style).
Thus, AIS forms a content-emotion-stylized image triplet problem, and we construct EmoStyleSet from ArtEmis~\cite{achlioptas2021artemis} to support it.
Stylized images and emotion labels come directly from ArtEmis.
To obtain content images, we adopt UnZipLoRA~\cite{liu2024unziplora} to disentangle content from style and use ControlNet~\cite{zhang2023adding} to generate realistic images.
Automatic and human evaluations are applied to filter noisy results.

We propose EmoStyle, an AIS framework capable of generating emotion-evoking, content-preserving images with appealing artistic styles, see Fig.~\ref{fig:teaser}.
EmoStyle consists of two main modules: \textit{Emotion-Content Reasoner} and \textit{Style Quantizer}.
In art, style and content are intertwined, with stylistic choices influenced by subject matter and emotion~\cite{gombrich1995story}.
Inspired by this, we develop an Emotion–Content Reasoner that integrates emotion and content cues through a transformer~\cite{kim2021vilt} to learn coherent style queries.
Unlike content, style often falls into discrete perceptual categories, \eg, \textit{Impressionism}, \textit{Modernism}, and \textit{Realism}.
We therefore introduce a Style Quantizer that discretizes continuous features into codebook entries, each representing a style prototype associated with a specific emotion.
This enables the Style Quantizer to align each emotion with multiple artistic styles, allowing for controllable and interpretable AIS.

To evaluate EmoStyle, we use the 405-image inference set following EmoEdit, given the same input format, \ie, emotion label and content image.
Beyond test images, our method applies to any user-provided content.
We compare EmoStyle with state-of-the-art style transfer, image editing, and AIM approaches.
Both quantitative and qualitative results confirm its effectiveness, further supported by a user study validating its alignment with human preferences. 
Additionally, the learned style dictionaries show strong adaptability to emotion-driven text-to-image generation tasks.

In summary, our contributions are:
\begin{itemize}
	\setlength{\itemsep}{0pt}
	\setlength{\parsep}{0pt}
	\setlength{\parskip}{0pt}
	
	\item We introduce Affective Image Stylization (AIS), which applies artistic styles to evoke target emotions while preserving its original content, and construct EmoStyleSet as its data foundation.
	
	\item We propose EmoStyle, which integrates an Emotion–Content Reasoner to fuse emotion and content and a Style Quantizer to map continuous features into discrete, emotion-aware style representations.
	
	\item We conduct extensive experiments and user studies, showing that EmoStyle achieves a strong balance between emotional expressiveness and content consistency. It also shows strong potential for affective art creation.
	
\end{itemize}
\section{Related work}
\label{sec:rw}

\subsection{Artwork Emotion Analysis}
As a powerful medium for expressing emotions, art has garnered increasing research interest in Artwork Emotion Analysis in recent years~\cite{achlioptas2021artemis, mohamed2022okay, bose2021understanding, ishikawa2023affective, chen2024exploring, zhang2025emoart}.
Leveraging the large-scale WikiArt dataset~\cite{saleh2015large}, ArtEmis~\cite{achlioptas2021artemis} and ArtEmis-V2~\cite{mohamed2022okay} enriched emotion-related attributes and employed a trained neural speaker to capture finer emotional nuances in artworks.
Joshi \etal~\cite{joshi2011aesthetics} analyzed the problems of aesthetics and emotions in images from a computational perspective and discussed future directions in this research domain.
Bose \etal~\cite{bose2021understanding} proposed a multimodal approach to artwork emotion recognition, incorporating emotion-related elements.
Ishikawa \etal~\cite{ishikawa2023affective} introduced the affective artwork captioning task, utilizing attention mechanisms to enhance emotional understanding.
Chen \etal~\cite{chen2024exploring} investigated local emotion stimuli and established a benchmark dataset with baseline evaluations.
Zhang \etal~\cite{zhang2025emoart} introduced EmoArt, a large-scale art dataset where each image is carefully annotated with fine-grained emotion labels.
Nevertheless, previous studies have primarily focused on recognizing and interpreting emotions in artworks rather than generating them.
With the rise of generative AI, we introduce Affective Image Stylization (AIS) to explore how artistic styles can be used to evoke visual emotions.

\subsection{Style Transfer}
The task of style transfer can be categorized into reference image-based style transfer \cite{gatys2016image, chen2021dualast, wang2023stylediffusion, chung2024style, hertz2024style, xing2024csgo, wu2025uso, wang2025omnistyle} and text-guided style transfer \cite{li2020manigan, kwon2022clipstyler, fu2022language, huang2024diffstyler, wang2025omnistyle}.
Among image-based methods, Wang \etal \cite{wang2023stylediffusion} introduced a new method to achieve disentangle content-style in the CLIP image space.
Chung \etal~\cite{chung2024style}, Hertz \etal~\cite{hertz2024style} and Xing \etal~\cite{xing2024csgo} refined the features of attention layers to generate style-consistent and content-preserving results.
Wu \etal~\cite{wu2025uso} presented a well-performed model via disentangled and reward learning.
In text-guided approaches, Kwon \etal~\cite{kwon2022clipstyler} proposed a framework that enables style transfer solely through text descriptions.
Fu \etal~\cite{fu2022language} introduced Contrastive Language-Visual Artist (CLVA), which learns to extract visual semantics from style instructions using a patch-wise style discriminator.
Zhang \etal~\cite{zhang2023inversion} treated style as a learnable textual description, effectively capturing and transferring artistic styles.
Wang \etal~\cite{wang2025omnistyle} accomplished both text-guided and image-based tasks by constructing a high-quality dataset.
While existing methods can produce aesthetically appealing results, they often rely on style images or art-knowledge prompts, which demands a certain level of expertise.
Emotions, in contrast, are universally understood and easy to specify.
To bridge this gap, we use emotion words as prompts to guide stylized generation and construct a dataset to support this task.

\subsection{Affective Image Manipulation}
Affective Image Manipulation (AIM) has evolved along two directions: style-based methods, which modify colors and styles to evoke emotions~\cite{yang2008automatic, liu2018emotional, sun2023msnet, weng2023affective}, and content-based methods, which incorporate emotional elements into images~\cite{yang2025emoedit, lin2025make}.
Early AIM research primarily focused on color manipulation to influence emotions.
Yang \etal~\cite{yang2008automatic} pioneered the use of color transfer for emotion-driven image editing, while Liu \etal~\cite{liu2018emotional} introduced an end-to-end color transfer approach.
More recently, AIM has expanded beyond color adjustments to incorporate emotional elements into image content.
Yang \etal~\cite{yang2025emoedit} introduced EmoEdit, building a task-specific dataset and training an Emotion Adapter to modify image content to express emotions.
Lin \etal~\cite{lin2025make} proposed EmoEditor, a human-preference aligned diffusion model, which evokes emotions by manipulating image regions.
Meanwhile, style-based AIM has evolved to explore emotion-aware stylization.
Sun \etal~\cite{sun2023msnet} developed MSNet, a deep architecture that employs a Multi-Sentiment Semantics Space for sentiment-aware style transfer.
Weng \etal~\cite{weng2023affective} introduced the Affective Image Filter (AIF), designing to translate textual descriptions into visual representations.
Despite significant progress in emotion-aware style transfer, existing methods often rely on reference images or carefully crafted textual descriptions, which limits their accessibility. 
Given an emotion word and a content image, EmoStyle introduces an Emotion–Content Reasoner to extract emotion-aware style query and a Style Quantizer to fetch its suitable prototype.
The style dictionary establishes an interpretable mapping between emotions and styles, ensuring the expressive generative results.
\section{Method}
\label{sec:method}

\subsection{EmoStyleSet}
\label{sec:emostyleset}

Existing artwork datasets (\eg, ArtEmis~\cite{achlioptas2021artemis}, EmoArt~\cite{zhang2025emoart}) serve as valuable benchmarks for artwork emotion analysis but remain inadequate for Affective Image Stylization (AIS).
These datasets assign emotion labels to artworks without distinguishing whether the emotion stems from \textit{content} or \textit{style}.
Since AIS aims to learn how style conveys emotion, separating style from overall artwork is essential.
To this end, we build EmoStyleSet, derived from ArtEmis, containing 10,041 triplets of content image, target emotion, and stylized image.

Specifically, ArtEmis is a large-scale visual emotion dataset containing 80K artworks, designed to capture the affective responses elicited by visual art.
Each image is annotated by multiple evaluators, resulting in an emotion distribution per artwork.
We convert these distributions into single labels and remove neutral samples to obtain simpler and accurate emotion annotations.
As both content and style influence emotional perception~\cite{mohammad2018wikiart, jia2012can}, we separate content from style to study how artistic styles evoke emotions.

To extract the content of each artwork, we employ BLIP-2~\cite{li2023blip} to generate descriptive captions for ArtEmis.
Given an artwork and its caption, UnZipLoRA~\cite{liu2024unziplora} decomposes the image into a content LoRA and a style LoRA.
We further convert each artwork into a Canny edge map to preserve structural information.
The Canny image and caption are then jointly input to ControlNet~\cite{zhang2023adding}, where the content LoRA is injected to maintain semantic consistency.
The output of ControlNet is regarded as the content image, with the style removed from the original ArtEmis artwork.

\begin{figure}
	\centering
	\includegraphics[width=\linewidth]{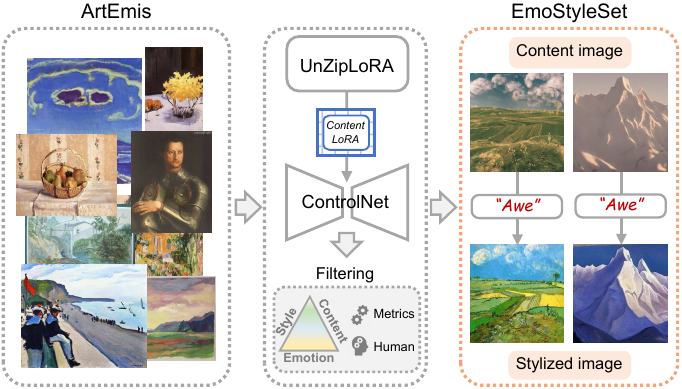}
	\vspace{-5pt}
	\caption{Construction process of EmoStyleSet. Given artworks from ArtEmis, after generation and filtering, each triplet contains a content image, a target emotion, and a stylized image.
	}
	%	\Description{}
	\label{fig:method_1}
	\vspace{-5pt}
\end{figure}

\begin{figure*}
	\centering
	\includegraphics[width=\linewidth]{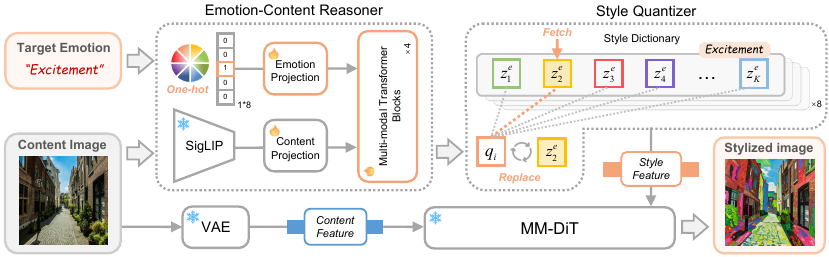}
	\vspace{-5pt}
	\caption{Overview of EmoStyle. We introduce an Emotion–Content Reasoner to integrate emotion and content features, and a Style Quantizer to map continuous queries to discrete style prototypes, generating stylized images with faithful emotion and preserved content. 
	}
	%	\Description{}
	\label{fig:method_2}
	\vspace{-5pt}
\end{figure*}

Since the content images in EmoStyleSet are generated in an unsupervised manner, certain inaccuracies may occur.
To ensure data quality, we perform filtering from three aspects: style, content and emotion, as illustrated in Fig.~\ref{fig:method_1}.
For content, we evaluate semantic consistency using CLIP similarity~\cite{radford2021learning} and measure structural preservation with LPIPS.
Emotion accuracy is assessed by a classifier trained on ArtEmis, ensuring each triplet matches its target emotion.
Finally, human annotators verify that the stylized images display clear stylistic differences from their corresponding content images.
More details are provided in the supplementary materials.

As shown in Fig.~\ref{fig:method_1}, EmoStyleSet is organized as \{content image, emotion word, stylized image\}, denoted as content-emotion-stylized image triplets, following the standard style transfer paradigm.
Taking \textit{awe} as an example, the upper row shows content images of landscapes and mountains, while the lower row presents their stylized counterparts that evoke a sense of magnificence and vastness.

\subsection{EmoStyle}

\subsubsection{Emotion-Content Reasoner}

In artistic creation, style and content are deeply intertwined, with artists naturally adapting their stylistic choices to align with the subject matter and emotional intent~\cite{gombrich1995story}.
Consequently, a key challenge in AIS is emotion-aware stylization, \ie, evoking emotions through appropriate artistic styles.
To address it, we first propose Emotion-Content Reasoner, designed to determine the most appropriate style given a content image and a target emotion.
As illustrated in Fig.~\ref{fig:method_2}, we employ SigLIP~\cite{zhai2023sigmoid} to extract semantic features from the content image.
Prior works~\cite{yang2024emogen, yang2025emoedit} modeled emotions as textual inputs, which is inadequate because LLMs often associate emotion words with facial expressions, whereas emotions in AIS involve a broader artistic context.
Therefore, we encode each emotion as a 1*8 one-hot vector, which offers two advantages: (1) emotions are mutually orthogonal; (2) all vectors collectively span the emotion space.
Since semantic and emotion features lie in different spaces, we project them through separate layers into a unified embedding space.
The fused embeddings are then processed by a four-block multi-modal transformer~\cite{kim2021vilt}, which applies both self- and cross-attention to model interactions between emotion and content features.
Through this interaction, the transformer reasons across modalities to derive an emotion-aware, content-conditioned style query $q_i$ for subsequent style selection:
\begin{equation}
	\label{eq:reasoner}
	\begin{aligned}
		\overline{q_i^{k}} &= MSA(LN(q_i^{k-1})) + q_i^{k-1},\\
		q_i^{k} &= MLP(LN(\overline{q_i^{k}})) + \overline{q_i^{k}},
	\end{aligned}
\end{equation}
where $LN$ denotes the LayerNorm, $k$ indexes the transformer blocks. 
Specifically, we initialize $q_i^0$ by concatenating the outputs of emotion and content projections.

\subsubsection{Style Quantizer}
\label{sec:codebook}

Artistic styles are often perceived categorically rather than continuously in human visual cognition, \eg, Impressionism, Modernism, Realism.
Motivated by this, we introduce a Style Quantizer to discretize style features to distinct prototypes, enabling interpretable and controllable style learning.
The relationship between emotion and style is inherently \textit{many-to-many}, as each emotion can be expressed through various styles, and each style can evoke multiple emotions. 
However, for the initial exploration in AIS, we adopt a \textit{one-to-many} simplification, following art history that emotions are conveyed through diverse styles~\cite{harrison2003art}.
%Under this assumption, we construct a style dictionary for each emotion, containing several representative styles.

In EmoStyleSet, each triplet is denoted as \{$I_c$, $e$, $I_s$\}, where $I_c$ represents content image, $e$ refers to emotion word, and $I_s$ is stylized image.
Inspired by VQ-VAE~\cite{van2017neural}, we construct eight style dictionaries, each corresponding to an emotion category, denoted as $Z^e = \{z_k^e\}_{k=1}^K$.
To be specific, we employ the style encoder ${\mathcal{E}_s(\cdot)}$ from USO~\cite{wu2025uso} to extract style features and compute pairwise similarities among all artworks.
A similarity threshold is then applied to initialize each style dictionary with distinct style prototypes.
In the first stage, we learn the prototype styles using Eq.~\ref{eq:style_loss}.
In the second stage, we apply vector quantization~\cite{van2017neural} $Q(\cdot)$ to the style query $q_i$ to fetch its nearest neighbor $z_k^e$ from style dictionary $Z^e$, thereby converting the representation from continuous to discrete:

\begin{align}
	\label{eq:codebook}
	Q(q_i) = z_k^{e}, \quad 
	\text{where } k = \underset{j}{\arg\min} \; \left\| q_i - z_j^{e} \right\|_2.
\end{align}

By adopting discrete style representations, we simplify style-emotion mapping and ensure coherence between style and content in artworks, effectively addressing the challenge outlined in the introduction.

\subsubsection{Training and Inference}
\label{sec:training}

As shown in Fig.~\ref{fig:method_2}, we adopt MM-DiT~\cite{wu2025uso} as the backbone.
Specifically, the content image is encoded by a VAE into a latent content feature, while the output of the Style Quantizer serves as the style feature.
These two features are then fed into MM-DiT for generation.
During training, we freeze the MM-DiT parameters and update only the Emotion-Content Reasoner and Style Quantizer.
Style Quantizer learns what each style prototype represents, while Emotion-Content Reasoner learns which prototype to select given an emotion–content pair.
Accordingly, EmoStyle adopts a two-stage training scheme: prototype learning followed by emotion–style alignment.

In the first stage, we learn the prototype style features $Z^e$ with the stylized images from EmoStyleSet:
\begin{equation}
	\label{eq:style_loss}
	\begin{aligned}
		\mathcal{L}_{style} &= \left\| z_k^{e} - \mathcal{E}_s(I_s) \right\|_2^{2},\\
		\text{where }\; k &= \underset{j}{\arg\min}\; \left\| \mathcal{E}_s(I_s) - z_j^{e} \right\|_2,
	\end{aligned}
\end{equation}
where $z_k^e$ is the $k$-th representation in the style dictionary for emotion $e$. 
This loss guides the style representations to form discrete and diverse prototypes distributed across the style space.
It can be viewed as a clustering process, where each style feature is assigned to its nearest cluster centroid, and the centroids are iteratively updated based on their assigned members.
As a result, each style dictionary contains representative style embeddings for each emotion, as visualized in Sec.~\ref{sec:exp}.

In the second stage, we use content-emotion-stylized image triplets to align the generated stylized results with the ground truth at both the feature and pixel levels.
To ensure high-quality generation and effectively guide pixel-level learning on paired data, we adopt the standard flow-matching loss:
\begin{align}
	\label{eq:diffusion_loss}
	{{\mathcal{L}}_{FM}}={\mathbb{E}}_{{x_0},t,\epsilon}[{w(t)}{\left\| {v_\theta}-{v_t} \right\|^{2}}],
\end{align}
where $v_\theta$ denotes the predicted velocity and $v_t = \frac{d\alpha_t}{dt}{x_0} + \frac{d\sigma_t}{dt}\epsilon$ the ground-truth velocity.

Moreover, for a given emotion and content image pair, to guide emotion-content reasoner to find its corresponding style representation, we propose an align loss:
\begin{align}
	\label{eq:align_loss}
	{{\mathcal{L}}_{align}}= \left\| q_i - z_k^{e} \right\|_2^{2},
\end{align}
where $z_k^{e}$ is the nearest neighbor of $q_i$, ensuring that the learned style queries follow the style distribution in EmoStyleSet.

Besides, to ensure emotion fidelity, we use the emotion score $e_n$ derived from the voting results in ArtEmis to weight the training losses of different samples:
\begin{equation}
	\label{eq:weighted_loss}
	\begin{aligned}
		\mathcal{L}_1 &= \frac{1}{N}\sum_{n} e_n \cdot \mathcal{L}_{style},\\[-0.3em]
		\mathcal{L}_2 &= \frac{1}{N}\sum_{n} e_n \cdot (\mathcal{L}_{FM} + \mathcal{L}_{align}),
	\end{aligned}
\end{equation}
where $e_n$ denotes the emotion score of the $n$-th sample and $N$ refers to the total number of training samples.

At inference, once the user provides a content image (\eg, a street view) and a target emotion (\eg, excitement), EmoStyle can generate a stylized image that is both emotion-evoking and aesthetically pleasing, as in Fig.~\ref{fig:method_2}.

\begin{figure*}
	\centering
 	\includegraphics[width=\linewidth]{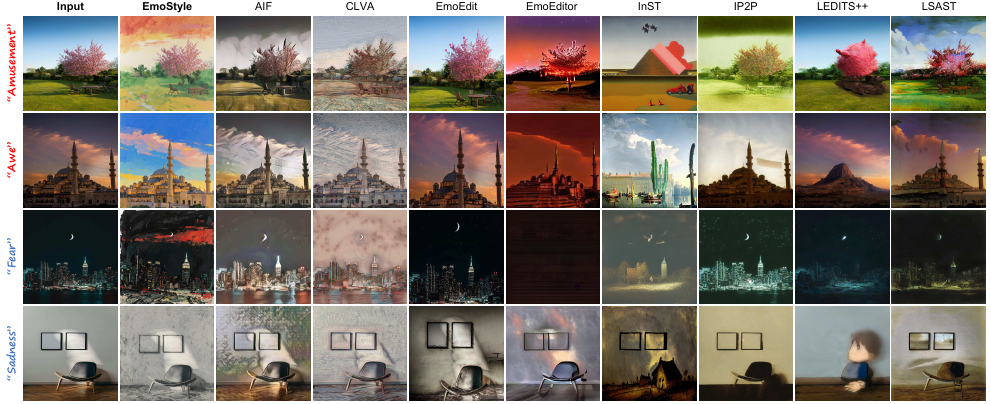}
	\vspace{-5pt}
	\caption{Comparison with the state-of-the-art methods, where EmoStyle surpasses others on emotion fidelity and aesthetic appeal.
	}
	%	\Description{}
	\label{fig:sota}
	\vspace{-5pt}
\end{figure*}
\section{Experiments}
\label{sec:exp}

\subsection{Dataset and Evaluation}

\paragraph{Dataset}

We train EmoStyle using content-emotion-stylized image triplets from the curated EmoStyleSet, as detailed in Sec.~\ref{sec:training}.
Since EmoStyle shares the same input format with EmoEdit~\cite{yang2025emoedit}, a content image and an emotion category, we evaluate it on the EmoEdit inference set, which contains 405 realistic, user-uploaded images collected from the internet.
Unlike EmoEdit, which produces realistic outputs, EmoStyle performs AIS and generates stylized results.
Each image is stylized into eight target emotions from Mikels’ wheel~\cite{mikels2005emotional}: \textit{amusement, awe, contentment, excitement, anger, disgust, fear}, and \textit{sadness}.
In total, our inference set contains 3,240 stylized results.

\paragraph{Evaluation Metrics}

Since AIS is a new task, we compile a set of metrics to evaluate it from multiple perspectives.
Like most style transfer methods, we first evaluate content preservation across different approaches.
Following OmniStyle~\cite{wang2025omnistyle}, we employ CLIP~\cite{radford2021learning} to assess semantic consistency between content and stylized images, and DINO~\cite{oquab2023dinov2} to measure pairwise structural preservation.
Unlike conventional style transfer methods, AIS focuses on emotional expressiveness in stylized images.
Thus, we introduce two emotion-related metrics.
Sentiment Gap (SG)~\cite{weng2023affective} measures how effectively the stylized images evoke the intended emotional responses.
Emotion Accuracy (Emo-A) assesses whether the stylized image aligns with the target emotion, using a pre-trained emotion classifier from ArtEmis~\cite{achlioptas2021artemis}.
In addition, Style Difference (SD)~\cite{deng2022stytr2} evaluates whether the stylized images exhibit appropriate color and textual consistency with the training data.
For more details, please refer to the supplementary materials.

\subsection{Comparisons}

\begin{table}
	\centering
	\scriptsize
	%	\small
	\caption{Comparisons with the state-of-the-art methods on style transfer, image editing and AIM methods. 
		%		Six metrics are used, encompassing pixel-level, semantic-level, and emotion-level evaluations.
	}
	\vspace{-5pt}
	\label{tab:exp_sota}
	\renewcommand\arraystretch{1}
	\setlength\tabcolsep{4.2pt}
	\begin{tabular}{lccccc}
		\toprule
		Method & CLIP $\uparrow$ & DINO $\uparrow$  & SG (\textperthousand) $\downarrow$ & Emo-A (\%) $\uparrow$ & SD $\downarrow$  \\
		\midrule
		LSAST~\cite{zhang2024towards} & 0.551 & 0.747 & 2.231 & 12.50 & 11.28 \\
		CLIPStyler~\cite{kwon2022clipstyler} & 0.709 & 0.769 & 3.001 & 12.60 & 19.89 \\
		InST~\cite{zhang2023inversion} & 0.569 & 0.679 & \underline{2.016} & 21.22 & 11.48 \\
		OmniStyle~\cite{wang2025omnistyle} & 0.710 & \underline{0.813} & 2.615 & 12.80 & 11.90 \\
		\midrule
		IP2P~\cite{brooks2023instructpix2pix} & 0.708 & 0.729 & 3.459 & \underline{24.34} & 12.76 \\
		LEDITS++~\cite{brack2024ledits++}  & 0.687 & 0.807 & 2.637 & 15.97 & 13.11  \\	
		\midrule
		EmoEditor~\cite{lin2025make} & 0.686 & 0.761 & 2.744 & 14.88 & 13.65 \\
		EmoEdit~\cite{yang2025emoedit} & 0.597 & 0.545 & 2.245 & 12.60 & 28.83 \\
		\midrule
		CLVA~\cite{fu2022language} & \textbf{0.727} & 0.789 & 2.030 & 14.99 & 9.49 \\
		AIF~\cite{weng2023affective} & 0.712 & 0.780 & 2.625 & 12.99 & \underline{8.48}  \\
		\midrule
		\textbf{EmoStyle} & \underline{0.718} & \textbf{0.842} & \textbf{1.976} & \textbf{33.36} & \textbf{7.59} \\
		\bottomrule
	\end{tabular}
	\vspace{-5pt}
\end{table}

To validate the effectiveness of EmoStyle, we compare it against the state-of-the-art methods across three categories:
(1) Style transfer: LSAST~\cite{zhang2024towards}, CLIPStyler~\cite{kwon2022clipstyler},  InST~\cite{zhang2023inversion}, and OmniStyle~\cite{wang2025omnistyle};
(2) Image editing: LEDITS++~\cite{brack2024ledits++}, and IP2P~\cite{brooks2023instructpix2pix};
(3) AIM: EmoEditor~\cite{lin2025make}, EmoEdit~\cite{yang2025emoedit}, CLVA~\cite{fu2022language}, and AIF~\cite{weng2023affective}.

\begin{figure}
	\centering
	\includegraphics[width=\linewidth]{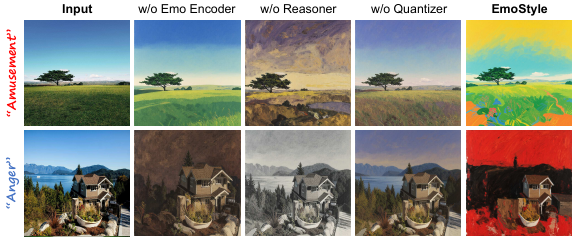}
	\vspace{-5pt}
	\caption{Ablation study. The Emotion Encoder, Emotion-Content Reasoner, and Style Quantizer are all shown to be effective.
	}
	%	\Description{}
	\label{fig:ablation}
	\vspace{-5pt}
\end{figure}

\paragraph{Quantitative Comparisons}

As shown in Table~\ref{tab:exp_sota}, EmoStyle outperforms other methods across most metrics, effectively evoking emotions while preserving content.
Unlike existing approaches that fail to capture emotional knowledge, EmoStyle achieves the highest Emo-A (33.36\%), surpassing the second-best by nearly 9\%, and records the lowest SG (1.976), owing to the effective interaction in the Emotion-Content Reasoner.
It also ranks first in DINO (0.842) and second in CLIP (0.718), demonstrating strong structural and semantic preservation with expressive styles.
While InST evokes emotions better than most methods, \ie, lower SG, it struggles to preserve semantic information.
The SD results further validate EmoStyle’s Style Quantizer, confirming its ability to learn accurate style representations from EmoStyleSet.
Overall, these results demonstrate EmoStyle’s superior balance between emotion expression and content preservation, making it an effective framework for emotion-driven artistic stylization.

\begin{figure}
	\centering
	\includegraphics[width=\linewidth]{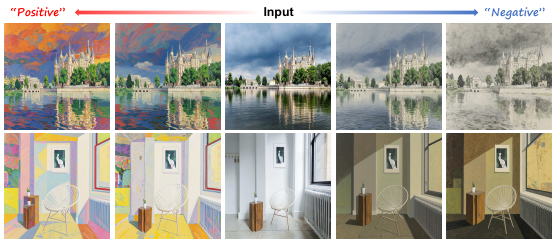}
	\vspace{-5pt}
	\caption{Ablation study on image guidance scale. EmoStyle can progressively edit an image towards different emotional polarities.
	}
	%	\Description{}
	\label{fig:guidance}
	\vspace{-5pt}
\end{figure}

\paragraph{Qualitative Comparisons}

We present the qualitative results in Fig.~\ref{fig:sota}.
Most compared methods struggle to convey emotions through stylization due to the lack of emotional knowledge.
Style transfer methods such as LSAST and InST can transform content images into artworks but fail to evoke the target emotion, resulting in visually appealing yet emotionally ambiguous results.
Among the two image editing methods, IP2P effectively introduces artistic elements and can evoke emotions to some extent after being trained on EmoStyleSet, highlighting the quality of the proposed dataset.
AIM methods such as EmoEdit and EmoEditor primarily modify colors or add emotional semantics to evoke emotions, but fail to achieve effective stylization.
Both CLVA and AIF can generate stylized images, but the style differences across emotions remain subtle.
By explicitly modeling the relationship between emotions and artistic styles, our approach produces stylized results that are both aesthetically appealing and emotionally evocative.

\begin{figure}
	\centering
	\includegraphics[width=\linewidth]{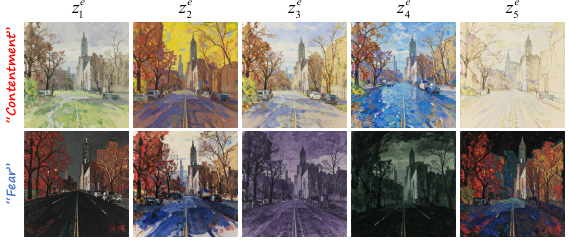}
	\vspace{-5pt}
	\caption{Visualization of the style dictionary, where each column corresponds to a distinct prototype $z_k^e$ for a specific emotion.
	}
	%	\Description{}
	\label{fig:vis}
	\vspace{-5pt}
\end{figure}

\begin{table}
	\centering
	\scriptsize
	%	\small
	\caption{User preference study. The numbers indicate the percentage of participants who vote for the result.}
	\vspace{-5pt}
	\label{tab:exp_userstudy}
	\renewcommand\arraystretch{1}
	\setlength\tabcolsep{3.8pt}
	\begin{tabular}{lcccccccc}
		\toprule
		Method & Aesthetic Perception $\uparrow$  & Emotion fidelity $\uparrow$ & Balance $\uparrow$ \\
		\midrule
		CLVA~\cite{fu2022language} & 8.50$\pm$12.13\% & 0.81$\pm$2.16\% & 1.19$\pm$7.07\%  \\
		InST~\cite{zhang2023inversion} & 2.50$\pm$4.86\% & 29.63$\pm$2.56\% & 1.34$\pm$5.53\%   \\
		AIF~\cite{weng2023affective}  & 9.08$\pm$9.42\% & 5.09$\pm$2.22\% & 7.76$\pm$9.91\%  \\
		\textbf{EmoStyle}  & \textbf{79.92$\pm$21.24\%}  & \textbf{64.47$\pm$4.55\%} & \textbf{89.70$\pm$14.48\%} \\
		\bottomrule
	\end{tabular}
	%	\vspace{-10pt}
\end{table}

\begin{figure*}
	\centering
	\includegraphics[width=\linewidth]{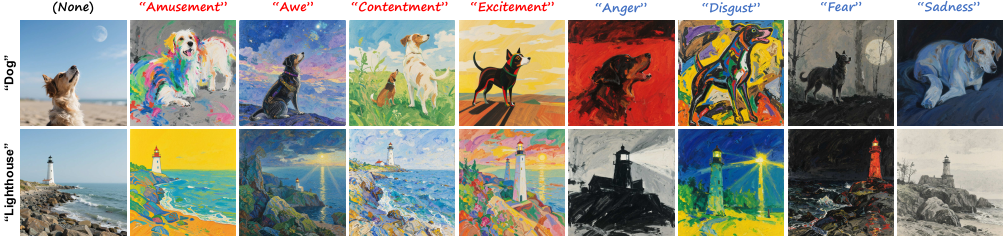}
	\vspace{-5pt}
	\caption{Emotion-driven text-to-image generation. Conditioned on a text prompt (content) and an emotion word (style), EmoStyle produces semantically faithful, emotionally expressive and aesthetically appealing stylized results.
	}
	%	\Description{}
	\label{fig:app}
	\vspace{-5pt}
\end{figure*}

\paragraph{User Study}
Since emotional perception is inherently abstract and subjective, we conducted a user study to assess human preferences for our method.
We invited 73 participants of varying ages and different backgrounds, and each evaluation session lasted approximately 15 minutes.
The study presented 24 sets of images, each containing one content image and four stylized versions generated by CLVA, InST, AIF and EmoStyle.
Participants were shown image sets and asked three questions regarding emotion, aesthetics, and their balance.
For each question, they selected one of four options, and we computed the vote percentages for each method.
As shown in Table~\ref{tab:exp_userstudy}, EmoStyle is the most preferred method across all evaluation aspects.
Balancing emotion and aesthetics is difficult, yet EmoStyle achieved the highest scores in both aspects, \ie, 64.47\% and 79.92\%, respectively.
Moreover, when evaluating the balance between emotion and aesthetics, 89.70\% of participants favored EmoStyle, demonstrating its strong alignment with human perception.
InST ranked the second in evoking emotions, scoring 29.63\%.
However, although it introduces emotional styles to images, it fails to preserve the original structure and semantics, as illustrated in Fig.~\ref{fig:sota}.

\subsection{Ablation Study}

\paragraph{Methodology}

In Fig.~\ref{fig:ablation}, we evaluate the effectiveness of the key components in EmoStyle, namely the Emotion Encoder, Emotion-Content Reasoner, and Style Quantizer.
Without the Style Quantizer, the results tend to appear overly realistic, indicating that the Emotion-Content Reasoner alone cannot map emotions to expressive artistic styles.
The Emotion Encoder and Emotion-Content Reasoner are both essential for emotion-aware stylization, as removing them leads to results that are less emotion-evoking and sometimes lack structural integrity.
Overall, through the integration of these key components, EmoStyle generates emotionally faithful, artistically appealing, and content-consistent images.

\paragraph{Guidance Scale}

We present EmoStyle's stylized results under varying image guidance scale in Fig.~\ref{fig:ablation}.
The middle image is the input, while the sides show stylization under varying guidance scales for positive and negative emotions.
Increasing the scale enhances emotion and style expressiveness but compromises structural preservation, revealing a trade-off between emotion and content.
However, EmoStyle effectively balances these aspects, as demonstrated in Table~\ref{tab:exp_sota}, Table~\ref{tab:exp_userstudy}, and Fig.~\ref{fig:sota}.
In addition, EmoStyle offers users flexible control over the image guidance scale, enabling adjustment of emotion intensity and fine-tuning of stylization according to personal preference.

\paragraph{Visualization}

As shown in Fig.~\ref{fig:ablation}, we visualize several style representations in the Style dictionary $Z^e = \{z_k^e\}_{k=1}^K$ for two distinct emotions.
Each style dictionary offers diverse and aesthetically appealing prototypes that effectively evoke the target emotion, reflecting the one-to-many emotion style mapping assumed in Sec.~\ref{sec:codebook}.
Users can further choose specific prototypes for each emotion to generate images that match their preferences.

\subsection{Applications}

\paragraph{Emotion-driven Text-to-Image Generation}

Beyond image stylization, EmoStyle can be extended to text-to-image generation, enabling the creation of emotionally expressive images from textual descriptions, as shown in Fig.~\ref{fig:app}.
Since SigLIP aligns image and text modalities, we replace the image encoder in Fig.~\ref{fig:method_2} with a text encoder for this task.
The Emotion-Content Reasoner then selects the most appropriate emotion-aware style based on the given textual description.
For example, given the prompt ``Dog'', our method generates eight stylized images, each conveying a target emotion while remaining artistically appealing.
These results highlight EmoStyle’s ability to preserve content integrity while ensuring emotion fidelity, expanding its potential for affective content creation and personalized design.
\section{Conclusion}
\label{sec:conclusion}

\paragraph{Discussion}

We present EmoStyle, a framework for Affective Image Stylization that applies artistic styles to evoke specific emotions while preserving content integrity.
Unlike conventional stylization methods that prioritize aesthetics, EmoStyle focuses on emotion-driven transformations.
The Emotion-Content Reasoner ensures content-consistent, emotion-aware stylization, while the Style Quantizer encodes multiple style prototypes for each emotion.
Experiments have demonstrated that EmoStyle effectively generates emotionally expressive and aesthetically appealing images.
As the initial attempt to explore the intrinsic relationship between emotion and style, EmoStyle offers greater flexibility and potential for broader applications in art and visual design.

\paragraph{Limitations}

Despite its advancements, EmoStyle still faces several challenges.
First, the mapping between emotion and style is inherently many-to-many.
Dark, muted tones may evoke melancholy in one image but mystery in another, while a single emotion such as contentment can be expressed through diverse artistic styles ranging from minimalist watercolor to detailed impressionism.
Second, emotion in artworks is evoked not only by style but also by content.
Understanding how the two visual stimuli interact to elicit emotions and developing methods that balance their contributions remain open research questions.
Finally, although our evaluation metrics capture several aspects, assessing the subjective emotional impact of stylized images is still challenging.
Future work should establish more comprehensive evaluation frameworks that incorporate human feedback and psychological insights.
{
    \small
    \bibliographystyle{ieeenat_fullname}
    \bibliography{EmoStyle}
}


\end{document}